# A New Self-Alignment Method without Solving Wahba Problem for SINS in Autonomous Vehicles

Hongliang Zhang*, Yilan Zhou, Lei Wang, and Tengchao Huang

*Abstract*—Initial alignment is one of the key technologies in strapdown inertial navigation system (SINS) to provide initial state information for vehicle attitude and navigation. For some situations, such as the attitude heading reference system, the position is not necessarily required or even available, then the self-alignment that does not rely on any external aid becomes very necessary. This study presents a new self-alignment method under swaying conditions, which can determine the latitude and attitude simultaneously by utilizing all observation vectors without solving the Wahba problem, and it is different from the existing methods. By constructing the dyadic tensor of each observation and reference vector itself, all equations related to observation and reference vectors are accumulated into one equation, where the latitude variable is extracted and solved according to the same eigenvalues of similar matrices on both sides of the equation, meanwhile the attitude is obtained by eigenvalue decomposition. Simulation and experiment tests verify the effectiveness of the proposed methods, and the alignment result is better than TRIAD in convergence speed and stability and comparable with OBA method in alignment accuracy with or without latitude. It is useful for guiding the design of initial alignment in autonomous vehicle applications.

## I. Introduction

THE initial alignment in strapdown inertial navigation system (SINS) is to determine the initial state of the navigation carrier, like attitude, velocity and position, and its accuracy and speed will have a great impact on subsequent navigation performance [1]. The initial velocity and position are usually provided by external sensors, whereas the initial attitude can be obtained by the system's inertial measurement unit (IMU) output or some external assistance [2]. For common situations where the carrier is not moving, such as moored ships, stopped vehicles, etc., the velocity is considered as zero and the position is unchanged, then the initial alignment can be seen as an attitude estimation problem, and it is usually solved by IMU data and the local latitude beforehand [3]. In some special applications, the accurate attitude information is needed while the position is not necessarily required, such as the gyroscope-stabilized platform, shipborne gyrocompass references, the communication satellite tracking, etc., once global navigation satellite system (GNSS) is ineffective to provide the latitude, like underground excavating machine [4] and underwater vehicles [5], the self-alignment method that does not rely on external aiding information is necessary and advantageous. Although many scholars have done lots of work on this topic of initial alignment, the latitude estimation and self-alignment technologies still face great challenges.

Based on the alignment state of the carrier, the initial alignment may occur under stationary, swaying or moving conditions. For the stationary alignment, an efficient way is based on some improved three-axis attitude determination (TRIAD) method [6], which can directly calculate the initial attitude of the carrier by using two non-collinear observation vectors, namely the earth rotation rate and local gravity vectors measured by IMU. However, to ensure its high alignment accuracy, providing a perfect static environment is a tough requirement in most practical cases [7]. For the swaying alignment, the earth rotation rate cannot be accurately measured by the gyroscope because of various external disturbances, which makes the swaying alignment more complicated than the stationary alignment. The swaying alignment methods are mainly based on the compass effect [8], filtering means [9], the idea of optimization [10]. Moreover, the inertial frame method that is based on the gravitational apparent motion (GAM) in the inertial frame [11] can isolate the swaying interference, and it is a popular means to achieve the initial attitude by using gravity observation vectors [12]. For the in-motion alignment, the system observability and vehicular maneuverability can be improved, but some external assistance is necessary for tracking the vehicle motion state [13]. In this work, the swaying self-alignment based on the GAM is further studied, and a new solution is provided for the alignment demands that the vehicle is under swaying and latitude uncertain conditions.

Under swaying conditions, the essence of initial alignment is a real-time attitude estimation problem. Owing to the conical movement of the local gravity in relation to the inertial space [14], i.e., GAM, the time-varying attitude estimation problem can be transformed into a constant attitude determination problem, which is then solved by using gravity observation vectors to determine the transformation matrix from a body reference frame to an inertial reference frame. Various methods have been developed to solve the problem based on observation vectors, and an excellent review can be found in Hashim [15]. As far as we know, one of the earliest and simplest attitude determination methods, known as the TRIAD algorithm, can be traced back to Black [16], and many of its modifications [6] are later proposed and serve as a doorway to the more advanced methods. Nonetheless, TRIAD methods still have an inherently shortcoming that it cannot make full use of all the observations and eventually fall into a suboptimal solution. Later, a least squares estimate based on minimizing all observation errors is defined as a Wahba problem [17], and to solve this problem as a common way to achieve the optimal attitude. Quaternion estimator (QUEST) and singular value decomposition (SVD) methods [18] are two classic and widely used methods, which are respectively based on the quaternion and the direction cosine matrix attitude representation, and many of the newly proposed methods are still based on solving the Wahba problem and their focus is on improving the efficiency of the solution, like fast linear attitude estimator [19]. Interestingly, Patera [20] proposed a new way to utilize all observation vectors by defining the inertia tensor of observation vectors, and this inertia tensor attitude estimation (VITAE) method does not solve the Wahba problem. This idea is meaningful, but it cannot ensure the uniqueness of the solution, thus failed in engineering applications. When these attitude estimation methods are applied in initial alignment for SINS, the local latitude needs to be given to calculate the local gravity and earth rotation rate vectors. Usually, the latitude is given by GNSS, when it is not available [21], determining the local latitude becomes an important issue.

*H. Zhang, Y. Zhou, L. Wang, and T. Huang are with College of Optical Science and Engineering, Zhejiang University, Hangzhou 310027, Zhejiang, China (email: zhanghl@zju.edu.cn).

To determine the latitude without external aids is meaningful. This problem originates from the stationary alignment with unknown latitude [22]. On the latitude determination under static conditions, Wang et al. [23] summarized four methods to calculate the latitude and their error equations were derived for analysis, and pointed out that the latitude error is mainly related to the IMU bias and position. Different from the stationary alignment, the latitude determination problem under swaying conditions becomes more complex due to the swaying interference, which invalidates these latitude determination methods under static conditions. For the swaying alignment, at present, two ways are mainly used to calculate the latitude without external aid: the first is based on the angle relationship between two vectors [24] and the other is based on the vector operation on three vectors [25], and noting that the used vectors are observation vectors of GAM at different moments. The former is a widely used method to determine the local latitude, and using this method means that the initial alignment is divided into two steps: i.e., the latitude determination and attitude estimation processes, which will increase the initial alignment time. For this reason, Li et al. [26] and Wang et al. [27] improved this method for higher accuracy and efficiency of initial alignment, and Yan et al. [28] provided a method to determine the positive-negative sign of the actual latitude, and the formulas of the latitude and attitude estimation errors is given for analysis. The latter can not only determine the latitude but also directly calculate the attitude matrix, which is a complete self-alignment method by constructing the earth axis vector [29]. Although the above two methods have been further improved, they are essentially based on the geometric relationship of two and three observation vectors, which means that the IMU data is not fully utilized, and the latitude and attitude estimation accuracy is not optimal.

This study is aimed at the swaying self-alignment problem with or without the local latitude information and provides a new solution that is different from the previous alignment ideas.

The proposed method is a new three axis attitude determination method (newTRIAD), which is inspired by VITAE [20] and similar to the classic TRIAD that provides an analytical solution. However, newTRIAD solves the problems that VITAE cannot provide a unique solution and TRIAD cannot take advantage of all observation vectors, and it is applied to solve the swaying alignment problem of SINS.

Moreover, the latitude estimation problem is also solved from the attitude estimation process of newTRIAD, then a new self-alignment of latitude and attitude determination (SALAD) method is proposed to solve the swaying alignment under geographic latitude uncertainty. The SALAD is completely different from the previous methods because it can make full use of all observation vectors to determine the latitude and attitude at the same estimation process.

Furthermore, the classic alignment methods based on TRIAD and optimization-based are used for comparing with newTRIAD and SALAD, and simulations, turntable swaying and ship mooring experiments are performed to verify the effectiveness of the proposed methods.

## II. METHODOLOGY

The right-handed Cartesian coordinate systems involved in this paper are defined as follows: $e$-frame is the common earth-centered earth-fixed coordinate system; $n$-frame is the navigation coordinate system, where the x-y-z axes coincide with the East-North-Up directions, respectively; $b$-frame is the body coordinate system, where the x-y-z sensitive axes of IMU coincide with the Right-Forth-Up directions, respectively; $i$-frame is the earth-centered inertial coordinate system. Moreover, $C_M^N$ denotes the attitude transformation matrix from the $M$-frame to the $N$-frame; $\omega_{AB}^C$ denotes the angular rate vector of the $B$-frame relative to the $A$-frame and projecting on the $C$-frame.

### A. Review of GAM-based swaying alignment

The special force function in $n$-frame for SINS gives

$$\begin{aligned}\dot{v}^n(t) &= f^n(t) - (2\omega_{ie}^n(t) + \omega_{en}^n(t)) \times v^n(t) + g^n \\ f^n(t) &= C_b^n(t)\tilde{f}^b(t) \\ g^n &= \begin{bmatrix}0 & 0 & -g\end{bmatrix}^T\end{aligned} \quad (1)$$

where $\dot{v}^n$ is the carrier acceleration; $f^n(t)$ is the special force vector that projects on $n$-frame, $C_b^n(t)$ is the body attitude matrix from $b$-frame to $n$-frame, $\tilde{f}^b(t)$ is the accelerometer triad output called special force vector; $\omega_{ie}^n$ is the earth's rotation angular velocity, $\omega_{en}^n$ is $n$-frame rotation angular velocity due to the carrier velocity; $g^n$ is the gravity vector and $g$ is the local gravity.

For the swaying base, the location of the carrier does not move significantly, then we can consider that $v^n(t)$ and $\dot{v}^n$ are equal to zero for simplification, and equation (1) is rewritten as

$$-g^n = C_b^n(t)\tilde{f}^b(t) \ . \tag{2}$$

Thus, the initial alignment task under swaying conditions is to obtain $C_b^n(t)$. Since $C_b^n(t)$ is time-varying and is hard to be estimated directly due to the output of the gyroscope and accelerometer triads with random swaying interference. The common solution [30] is to decompose $C_b^n(t)$ into three parts by using the chain rule as

$$C_b^n(t) = C_{n_0}^n(t) C_{b_0}^{n_0} C_b^{b_0}(t) \ , \tag{3}$$

where $n_0$-frame and $b_0$-frame are the navigation inertial coordinate system and the body inertial coordinate system, which are coincide with the $n$-frame and the $b$-frame at the beginning of initial alignment, respectively, and they are not change relative to the inertial space throughout the alignment process; then substituting (3) into (2) can obtain

$$C_n^{n_0}(t)(-g^n) = C_{b_0}^{n_0} C_b^{b_0}(t) \tilde{f}^b(t) \tag{4}$$

where $C_{n_0}^n(t)$ and $C_b^{b_0}(t)$ are named as the $n$-frame rotation matrix and the $b$-frame swaying matrix with respect to the inertial space, respectively, their differential equations are

$$\begin{aligned}\dot{C}_n^{n_0}(t) &= C_n^{n_0}(t)[\omega_{in}^n(t)\times] \\ \dot{C}_b^{b_0}(t) &= C_b^{b_0}(t)[\tilde{\omega}_{ib}^b(t)\times]\end{aligned} \ , \tag{5}$$

where $[\tilde{\omega}_{in}^n(t)\times]$ and $[\tilde{\omega}_{ib}^b(t)\times]$ are the skew symmetric matrix of $\omega_{in}^n(t)$ and $\tilde{\omega}_{ib}^b(t)$, respectively; $\omega_{in}^n(t) = \omega_{ie}^n(t) + \omega_{en}^n(t) \approx \omega_{ie}^n$ due to zero speed of the carrier, and $\omega_{ie}^n = [0 \ \ \omega_{ie}\cos L \ \ \omega_{ie}\sin L]^T$ which means the local latitude $L$ must be given; $\tilde{\omega}_{ib}^b(t)$ is the real-time gyroscope output; the initial value of $C_n^{n_0}(t)$ and $C_b^{b_0}(t)$ at the $t_0$ moment are the identity matrix.

In (4), $C_{b_0}^{n_0}$ denotes the constant attitude matrix from $b_0$-frame and $n_0$-frame, and it is the initial value of $C_b^n(t)$ at the beginning of initial alignment. Solving $C_{b_0}^{n_0}$ to isolate the angular vibration interference is usually based on the observation vector of GAM, and the attitude determination model in swaying alignment can be represented as

$$f^{n_0}(t)/|f^{n_0}(t)| = C_{b_0}^{n_0} f^{b_0}(t)/|f^{b_0}(t)| \ , \tag{6}$$

where $f^{b_0}(t)$ and $f^{n_0}(t)$ are the observation vector and the reference vector, respectively, and expressed as

$$\begin{aligned}f^{b_0}(t) &= C_b^{b_0}(t)\tilde{f}^b(t) \\ f^{n_0}(t) &= C_n^{n_0}(t)(-g^n)\end{aligned} \ , \tag{7}$$

where $\tilde{f}^b(t)$ is the real-time accelerometer output.

To smooth the measurement noise of $\tilde{f}^b(t)$, the integral form of (6) is usually used as

$$V^{n_0}(t) = C_{b_0}^{n_0} V^{b_0}(t) \ , \tag{8}$$

where $V^{n_0}(t)$ and $V^{b_0}(t)$ are named as velocity vectors as

$$\begin{aligned}V^{n_0}(t) &= \int_0^t C_b^{b_0}(\tau)\tilde{f}^b(\tau)d\tau \\ V^{b_0}(t) &= \int_0^t C_n^{n_0}(\tau)(-g^n)d\tau\end{aligned} \ . \tag{9}$$

So far, how to solve $C_{b_0}^{n_0}$ with high accuracy and efficiency in (8) is still a hot research topic. The commonly used two approaches are based on dual-vector attitude determination method, like TRAID, and multi-vector attitude determination method by solving the Wahba problem, such as OBA.

From (8), two velocity vector groups are extracted at two different moments $t_1$ and $t_2$, and usually $t_1 = t_2/2$, then the analytical solution of $C_{b_0}^{n_0}$ by TRAID gives

$$C_{b_0}^{n_0} = U^{n_0}[U^{b_0}]^{-1} = U^{n_0}[U^{b_0}]^T \ , \tag{10}$$

where $U^{n_0}$ and $U^{b_0}$ are orthogonal matrices, which are constructed by two non-colinear reference vectors and the corresponding observation vectors, respectively, as

$$\begin{aligned}U^{n_0} &= \begin{bmatrix} [V^{n_0}(t_1)]^T \\ [V^{n_0}(t_1) \times V^{n_0}(t_2)]^T \\ [V^{n_0}(t_1) \times V^{n_0}(t_2) \times V^{n_0}(t_1)]^T \end{bmatrix}^T \\ U^{b_0} &= \begin{bmatrix} [V^{b_0}(t_1)]^T \\ [V^{b_0}(t_1) \times V^{b_0}(t_2)]^T \\ [V^{b_0}(t_1) \times V^{b_0}(t_2) \times V^{b_0}(t_1)]^T \end{bmatrix}^T\end{aligned} \ , \tag{11}$$

where the superscript T denotes the matrix transpose.

Another way to obtain $C_{b_0}^{n_0}$ is to minimize the Wahba's loss function based on lots of velocity vector groups as

$$\min F(C_{b_0}^{n_0}) = \frac{1}{2}\sum_{k=1}^n w_k |V^{n_0}(t_k) - C_{b_0}^{n_0} V^{b_0}(t_k)|^2 \ , \tag{12}$$

where $w_k$ is the weight of measurement errors for the $k$ th sampling. Generally, $C_{b_0}^{n_0}$ is treated as an optimization variable that can be solved by SVD or be represented by a quaternion solved by QUEST and some other optimization-based methods.

When $C_{b_0}^{n_0}$ is obtained, the real-time $C_b^n(t)$ can be achieved by (3) and the swaying alignment is finished.

*B. Swaying alignment by a newTRIAD method*

The proposed newTRIAD is also based on GAM, but it is different from the above two kinds of approaches to achieve $C_{b_0}^{n_0}$. The method description can be continued from (8) and as follows.

Multiplying both sides of (8) by their transpose gives

$$\begin{aligned}V^{n_0}(t)[V^{n_0}(t)]^T &= C_{b_0}^{n_0} V^{b_0}(t)[C_{b_0}^{n_0} V^{b_0}(t)]^T \\ &= C_{b_0}^{n_0}[V^{b_0}(t)V^{b_0}(t)^T][C_{b_0}^{n_0}]^T\end{aligned} \ , \tag{13}$$

and the constructed dyadic tensors of the reference and observation vectors themselves are expressed as

$$\begin{aligned}T_r(t) &= V^{n_0}(t)[V^{n_0}(t)]^T \\ T_w(t) &= V^{b_0}(t)[V^{b_0}(t)]^T\end{aligned} \ , \tag{14}$$

then, equation (13) is rewritten as

$$T_r(t) = C_{b_0}^{n_0} T_w(t) [C_{b_0}^{n_0}]^T \ , \tag{15}$$

where the dyadic tensors $T_r(t)$ and $T_w(t)$ are symmetric $3\times 3$ matrices and their ranks equal 1. Since $C_{b_0}^{n_0}$ is orthonormal matrix, i.e., $[C_{b_0}^{n_0}]^T = [C_{b_0}^{n_0}]^{-1}$, then $T_r(t)$ and $T_w(t)$ are similar matrices and they have the same real eigenvalues.

The next deduction is based on (15) for achieving $C_{b_0}^{n_0}$. For a set of observation vectors, their total corresponding dyadic tensors satisfy

$$\begin{aligned}\sum_{k=1}^n w_k T_r(t_k) &= \sum_{k=1}^n w_k \left[C_{b_0}^{n_0} T_w(t_k)[C_{b_0}^{n_0}]^{-1}\right] \\ &= C_{b_0}^{n_0}\left[\sum_{k=1}^n w_k T_w(t_k)\right][C_{b_0}^{n_0}]^{-1}, \ n \geq 2\end{aligned} \ , \tag{16}$$

where $w_k$ is the weight of the $k$ th dyadic tensor, and it is related to the measurement error of the corresponding observation vector. Generally, the sum of all weights is equal to 1, and here, $w_k$ is set to $1/n$, which means every observation vector has an equal measurement error.

Next, equation (16) is simplified as

$$T_r = C_{b_0}^{n_0} T_w [C_{b_0}^{n_0}]^{-1} \ , \tag{17}$$

where
$$T_r = \sum_{k=1}^{n} w_k T_r(t_k)$$
$$T_w = \sum_{k=1}^{n} w_k T_w(t_k)$$
(18)

Similarly, $T_r$ and $T_w$ are similar and they can be easily diagonalized with the same diagonal elements $\lambda_1$, $\lambda_2$, $\lambda_2$, and suppose their diagonal matrices are $\Lambda$, and
$$\Lambda = \mathrm{diag}(\lambda_1 \quad \lambda_2 \quad \lambda_3), \quad \lambda_1 \geq \lambda_2 \geq \lambda_3 \geq 0$$
(19)

then, equation (17) is expanded as
$$U_r \Lambda U_r^{-1} = C_{b_0}^{n_0} U_w \Lambda U_w^{-1} \left[C_{b_0}^{n_0}\right]^{-1}$$
$$= C_{b_0}^{n_0} U_w \Lambda \left[C_{b_0}^{n_0} U_w\right]^{-1}$$
(20)

and let $U_r = C_{b_0}^{n_0} U_w$ to satisfy (20), then $C_{b_0}^{n_0}$ can be expressed as
$$C_{b_0}^{n_0} = U_r [U_w]^{-1} = U_r [U_w]^{\mathrm{T}}$$
(21)

where $U_r$ and $U_w$ are the orthogonal transformation matrices that transform $\Lambda$ to $T_r$ and $T_w$, respectively, and their column vectors are the eigenvectors corresponding to the different eigenvalues of $T_r$ and $T_w$ as
$$U_r = [X_{r,1} \quad X_{r,2} \quad X_{r,3}]$$
$$U_w = [X_{w,1} \quad X_{w,2} \quad X_{w,3}]$$
(22)

and the eigenvectors can be obtained by the eigenvalue decomposition of $T_r$ and $T_w$, that is
$$\begin{cases} T_r X_{r,j} = \lambda_j X_{r,j} \\ T_w X_{w,j} = \lambda_j X_{w,j} \end{cases}, j = 1, 2, 3.$$
(23)

Note that (21) cannot give a unique solution of $C_{b_0}^{n_0}$, because $U_r$ and $U_w$ are not unique due to the direction uncertainty of the eigenvector, and this problem is ignored in the original VITAE. For instance, both $X_{r,1}$ and $-X_{r,1}$ satisfy (23), and which are the eigenvectors corresponding to the eigenvalue $\lambda_1$ of $T_r$, then $U_r$ has $2^3$ kinds of combinations, and so do $U_w$. In order to solve this problem, some constraints are introduced to guarantee the direction consistency of their column vectors of $U_r$ and $U_w$, and the uniqueness and exactness of $C_{b_0}^{n_0}$.

Put (8), (21) and (22) together, we have
$$V^{n_0}(t) = C_{b_0}^{n_0} V^{b_0}(t)$$
$$X_{r,j} = C_{b_0}^{n_0} X_{w,j}, j = 1, 2, 3,$$
(24)

then, the eigenvectors $X_{r,j}$ and $X_{w,j}$ satisfy
$$V^{n_0}(t)^{\mathrm{T}} X_{r,j} = V^{b_0}(t)^{\mathrm{T}} X_{w,j}, j = 1, 2, 3,$$
(25)

Because of the observation noise, both sides of (25) will not be absolutely equal, thus we replace (25) with a robust from as
$$\left(V^{n_0}(t)^{\mathrm{T}} X_{r,j}\right)\left(V^{b_0}(t)^{\mathrm{T}} X_{w,j}\right) > 0, j = 1, 2, 3,$$
(26)

where $V^{n_0}(t)$ and $V^{b_0}(t)$ are not orthogonal with $X_{r,j}$ and $X_{w,j}$, respectively, and usually taken from the last sample for each attitude matrix update. Generally, equation (26) is used to ensure the uniqueness of (21), and it is implemented as follows,
$$X_{r,j} = \begin{cases} -X_{r,j}, & \text{if } \left(V^{n_0}(t)^{\mathrm{T}} X_{r,j}\right)\left(V^{b_0}(t)^{\mathrm{T}} X_{w,j}\right) < 0 \\ X_{r,j}, & \text{otherwize} \end{cases}, j = 1, 2, 3$$
or $X_{r,3} = X_{r,1} \times X_{r,2}$, $X_{w,3} = X_{w,1} \times X_{w,2}$
(27)

where $X_{r,3}$ and $X_{w,3}$ can be determined by $X_{r,1}$, $X_{r,2}$ and $X_{r,1}$, $X_{r,2}$, respectively, through the right-hand rule. That is, if the condition of (26) is not met, then change the direction of eigenvector $X_{r,j}$ to guarantee the direction consistency of $X_{r,j}$ and $X_{w,j}$. Of cause, changing the direction of $X_{w,j}$ is the same.

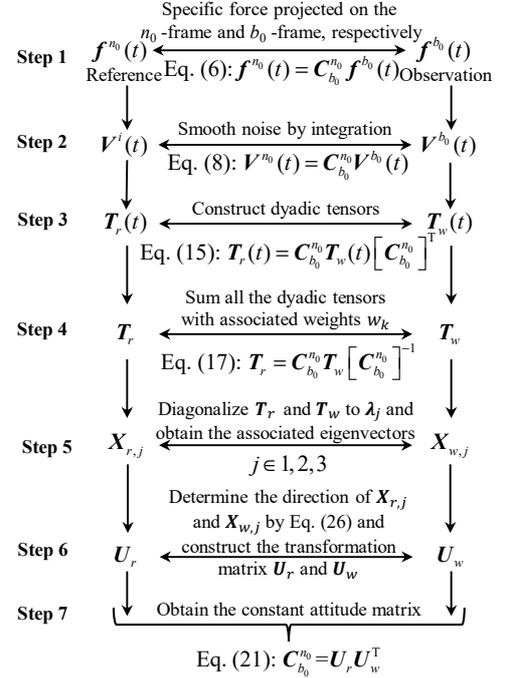

Fig. 1. Flowchart of newTRIAD in swaying alignment with latitude

Finally, put the obtained $C_{b_0}^{n_0}$ into (3) to complete the swaying alignment. For a clear description, the flowchart of newTRIAD is given in Fig. 1.

The proposed newTRIAD is compared with the classic TRIAD and solving Wahba problem methods from three aspects, which are the expression of solving $C_{b_0}^{n_0}$, the number of using observation vectors, and the solution properties. As shown in Table 1, newTRIAD and TRIAD seems to have similar solution expressions, but the construction of their orthogonal matrices $U$ by observation and reference vectors are quite different, because TRIAD just use two non-collinear vectors, while newTRIAD can utilize as many vectors as possible by performing a simple construction process (13), which has the advantage of smoothing the observation noise quickly and ensuring the accuracy and stability of alignment results. Although some optimized

In addition, newTRIAD mainly involves the eigenvalue decomposition of two symmetric 3×3 matrices, which is easy to implement and does not involve iterations compared with some OBA methods.

As a result, newTRIAD has the main advantages of TRIAD and OBA methods, namely, the analytical property and utilizing all observation vectors, but it does not need to solve the Wahba problem, and it is successfully applied to solve initial alignment problems for SINS.

TABLE 1
COMPARISON OF THREE KINDS OF ALIGNMENT METHODS

| method | TRIAD | optimization-based | newTRIAD |
|---|---|---|---|
| expression | $C_{b_0}^{n_0} = U^{n_0}\left[U^{b_0}\right]^{\mathrm{T}}$ | $\min F(C_{b_0}^{n_0})$ | $C_{b_0}^{n_0} = U_r [U_w]^{\mathrm{T}}$ |
| vectors | 2 | > 2 | > 2 |
| property | analytic | optimized | analytic |

## C. Self-alignment of latitude and attitude determination

In addition to the advantages of efficient analytical features and utilizing all observation vectors, the newTRIAD method can be modified to estimate the latitude and attitude at the same calculation process for solving the self-alignment problem under geographic latitude uncertainty, namely the proposed SALAD. Because SALAD can make use of all observation vectors, it is conductive to improving the accuracy and stability of latitude estimation compared with the currently used methods with partial observation vectors. SALAD is derived as follows.

Owing to the latitude is just related to the navigation frame, then in (3), $C_{n_0}^n(t)$ and $C_{b_0}^{n_0}$ include the latitude information, thus it is necessary to extract the latitude $L$ from both $C_{n_0}^n(t)$ and $C_{b_0}^{n_0}$. To facilitate the extraction of $L$, equation (3) can also be rewritten as

$$C_b^n(t) = C_{e'}^n C_{i_0}^{e'}(t) C_{b_0}^{i_0} C_b^{b_0}(t) , \qquad (28)$$

where $i_0$-frame is the special earth-centered inertial coordinate system, which is coincide with the $e'$-frame that is the $e$-frame rotates counterclockwise around its z-axis by the value of the local longitude $\lambda$, at the beginning of initial alignment, then $C_{e'}^{n_0}$ is related to $L$ as

$$C_{e'}^{n_0} = \begin{bmatrix} 0 & 1 & 0 \\ -\sin L & 0 & \cos L \\ \cos L & 0 & \sin L \end{bmatrix} , \qquad (29)$$

$C_{i_0}^{e'}(t)$ is updated in real time by the earth rate $\omega_e$ as

$$C_{i_0}^{e'}(t) = \begin{bmatrix} \cos(\omega_e t) & \sin(\omega_e t) & 0 \\ -\sin(\omega_e t) & \cos(\omega_e t) & 0 \\ 0 & 0 & 1 \end{bmatrix} . \qquad (30)$$

Submitting (28) to (2) can obtain

$$\begin{cases} f^{i_0}(t) = C_{b_0}^{i_0} f^{b_0}(t) \\ f^{i_0}(t) = C_{e'}^{i_0} C_n^{e'}(t)(-g^n) \\ f^{b_0}(t) = C_b^{b_0}(t) \tilde{f}^b(t) \end{cases} , \qquad (31)$$

Because $C_{e'}^{i_0}$, $C_n^{e'}(t)$ and $g^n$ are known, then equation (31) can be simplified as

$$\begin{bmatrix} g\cos L \cos(\omega_e t) \\ g\cos L \sin(\omega_e t) \\ g\sin L \end{bmatrix} = C_{b_0}^{i} C_b^{b_0}(t) \tilde{f}^b(t) . \qquad (32)$$

Since $g$ is related to $L$, their complex relational model will make latitude estimation difficult. Actually, $g$ is eliminated by normalizing both sides of (32), which means that the solution of the orthonormal matrix $C_{b_0}^{i_0}$ has nothing to do with the size of $g$. Normalizing (32) gives

$$S f_N^{i_0}(t) = C_{b_0}^{i_0} f_N^{b_0}(t) , \qquad (33)$$

and

$$\begin{aligned} S &= \text{diag}(\cos L \ \cos L \ \sin L) \\ f_N^{i_0}(t) &= [\cos(\omega_e t) \ \sin(\omega_e t) \ 1]^T \\ f_N^{b_0}(t) &= C_b^{b_0}(t) \tilde{f}^b(t) / |C_b^{b_0}(t) \tilde{f}^b(t)| \end{aligned} , \qquad (34)$$

where the subscript N denotes the normalization form of the reference and observation vectors, and $L$ is extracted into $S$ that is only related to $L$. For swaying base, $L$ can be a constant value, then as the constant coefficient matrix of $f_N^{i_0}(t)$, $S$ does not affect the use of the newTRIAD to derive $C_{b_0}^{i_0}$. The integral forms of $f_N^{i_0}(t)$ and $f_N^{b_0}(t)$ in (33) are represented by $V_N^{i_0}(t)$ and $V_N^{b_0}(t)$, respectively, then refer to (13) and (16), constructing their dyadic tensors and summing them up gives

$$S T_{N,r} S^T = C_{b_0}^{i_0} T_{N,w} [C_{b_0}^{i_0}]^{-1} , \qquad (35)$$

where $T_{N,r}$ and $T_{N,w}$ can be calculated as

$$\begin{aligned} T_{N,r} &= \sum_{k=1}^n w_k V_N^{i_0}(t_k) [V_N^{i_0}(t_k)]^T \\ T_{N,w} &= \sum_{k=1}^n w_k V_N^{b_0}(t_k) [V_N^{b_0}(t_k)]^T \end{aligned} . \qquad (36)$$

Similarly, $S T_{N,r} S^T$ and $T_{N,w}$ are similar matrices, and they have the same traces as

$$\text{tr}(S T_{N,r} S^T) = \text{tr}(T_{N,w}) , \qquad (37)$$

where tr denotes the trace of a matrix. Since $T_{N,r}$ and $T_{N,w}$ are known and $S$ contains only $L$, then $L$ can be solved and expressed as

$$L = \pm \text{acos}(\sqrt{\frac{\text{tr}(T_{N,w}) - T_{N,r33}}{T_{N,r11} + T_{N,r22} - T_{N,r33}}}) , \qquad (38)$$

where $T_{N,r11}$, $T_{N,r22}$, $T_{N,r33}$ are the diagonal elements of $T_{N,r}$ from the top-left to bottom-right corners, respectively; $\text{tr}(T_{N,w})$ is the sum of the diagonal elements of $T_{N,w}$. Note that the square root result in (38) may be greater than 1 due to IMU errors and when the latitude approaches zero, thus this constraint of less than 1 should be handled to meet the arc-cosine function. Moreover, the north latitude is indicated by a positive sign, and the south latitude is represented by a negative sign.

To determine of the sign of $L$, it depends on the attitude matrix $C_{b_0}^{i_0 \prime}$ that is calculated by using the positive latitude $L^+$, i.e., the absolute value of $L$ obtained by (38).

When $L$ is positive, equation (33) is expressed as

$$\begin{aligned} S^+ f_N^{i_0}(t) &= C_{b_0}^{i_0} f_N^{b_0}(t) \\ S^+ &= \text{diag}(\cos L^+ \ \cos L^+ \ \sin L^+) \end{aligned} , \qquad (39)$$

then $C_{b_0}^{i_0 \prime} = C_{b_0}^{i_0}$, and $C_{b_0}^{i_0} = U_r U_w^T$ by continuing to do the eigenvalue decomposition of (35) as used in newTRIAD method. Since $C_{b_0}^{i_0}$ is the right-hand coordinate orthonormal transformation matrix, then $\det(C_{b_0}^{i_0})=1$, i.e., $\det(C_{b_0}^{i_0\prime})=1$.

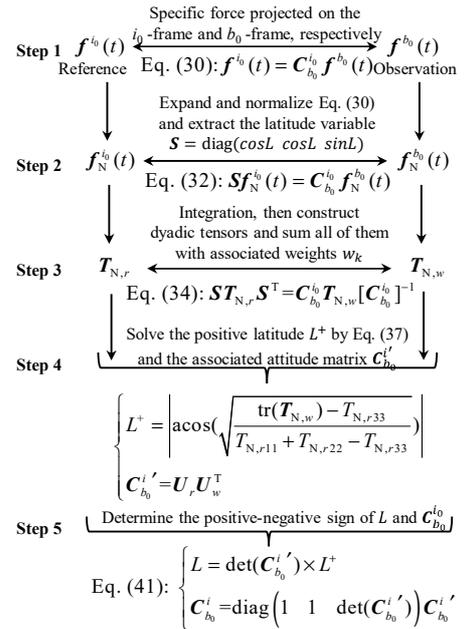

Fig. 2. Flowchart of SALAD in swaying alignment without latitude.

When $L$ is negative, equation (33) is expressed as

$$NS^+ f_N^{i_0}(t) = C_{b_0}^{i_0} f_N^{b_0}(t)$$
$$N = \text{diag}(1 \quad 1 \quad -1) \quad , \quad (40)$$
$$S^+ = \text{diag}(\cos L^+ \quad \cos L^+ \quad \sin L^+)$$

and equation (40) can be rewritten as

$$S^+ f_N^{i_0}(t) = N C_{b_0}^{i_0} f_N^{b_0}(t) \quad , \quad (41)$$

then we can get that $C_{b_0}^{i_0\prime} = N C_{b_0}^{i_0} = U_r U_w^T$, and $\det(C_{b_0}^{i_0\prime}) = -1$.

Based on the above analysis, the positive-negative sign of the latitude can be determined by the value of $C_{b_0}^{i_0\prime}$, then the local latitude and the constant attitude matrix are expressed as

$$L = \det(C_{b_0}^{i_0\prime}) \times L^+$$
$$C_{b_0}^{i_0} = \text{diag}(1 \quad 1 \quad \det(C_{b_0}^{i_0\prime})) C_{b_0}^{i_0\prime} \quad . \quad (42)$$

Finally, put the obtained $C_{b_0}^{i_0}$ into (28) to finish the swaying self-alignment without any external aid. The flowchart of the proposed SALAD is shown in Fig. 2.

## III. SIMULATION AND EXPERIMENT ANALYSIS

Simulation and experiment tests are carried out to demonstrate the effectiveness of the proposed newTRIAD and SALAD methods in solving the initial alignment problem with or without the local latitude. For the convenience of verification, the simulation and experiment results are compared with the classic GAM-based approaches, which include the classic dual-vector method, like TRIAD shown in (10), and the popular multi-vector methods, like optimization-based algorithm (OBA) by solving the Wahba problem in (12). Except for the core algorithms of different methods, all the conditions are kept consistent, and their attitude outputs and time costs are also recorded as an intuitive reference for comparison. Moreover, the simulation program and the experimental data processing are performed on the MATLAB platform because of its convenience.

### A. Simulation tests

The inertial sensor error is simplified into two terms, i.e., the constant bias and the random walk noise, and their installation errors are not considered. Table 2 lists the error parameters of gyroscopes and accelerometers in simulation. The real-time swaying attitude of IMU is described by Euler angles, i.e., pitch ($\theta$), roll ($\gamma$), yaw ($\psi$), which are some regular trajectories and simulated by three different cosine functions, $A\cos(2\pi/T \cdot t)$, where their amplitudes $A$ and periods $T$ are set to 7°, 10°, 5° and 5 s, 6 s, 7 s, respectively. Considering that this is a basic verification of initial alignment, the linear vibration interference is not introduced. Moreover, the default IMU output frequency is set to 50 Hz, the cosine swaying center, i.e., the initial attitude, is set to [0 0 0]$^T$, and the local latitude is 30.266°, then the corresponding misalignment angle limit is about -0.0057°, 0.0057°, -0.0882° in theory and used as an alignment reference.

TABLE 2
ERROR PARAMETERS FOR INERTIAL SENSORS IN SIMULATION

| Error term | Gyroscope | Accelerometer |
|---|---|---|
| Constant bias | 0.02 °/h | 100 μg |
| Random walk | 0.002 °/sqrt (h) | 10 μg/sqrt (Hz) |

1) Latitude estimation performance of SALAD

The estimated latitude in SALAD is based on all the observation vectors, here this latitude estimation method is named multi-vector latitude determination (MvLD). Two classical methods of latitude estimation in swaying base are used for comparation, the first is based on the angle relationship between two sets of observation vectors[28], which is named 2-vector latitude determination (2vLD); the other is based on the vector operation of three reconstructed observation vectors[25], which is named 3-vector latitude determination (3vLD). The latitude estimation error curves of these three methods are given in Fig. 3.

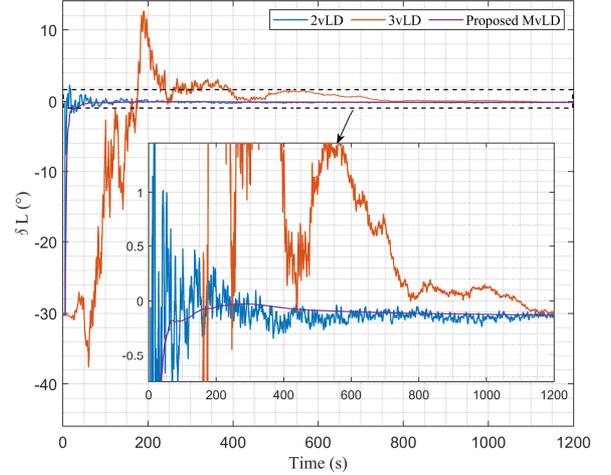

Fig. 3 Curves of latitude estimation error

As shown in Fig. 3, the latitude error curve of the proposed MvLD is much smoother than the classic 2vLD and 3vLD methods, this is mainly because MvLD makes full use of all observation vectors, and the observation noise is effectively suppressed. Although 2vLD can obtain two sets of vectors by smoothing for a short period of time, the noise reduction effect is relatively limited. The performance of 3vLD is relatively poor because it is necessary to reduce the observation noise based on the method of reconstructing the observation vector, while the reconstruction speed is relatively slow, and the reconstruction accuracy directly affects the latitude estimation accuracy. Moreover, MvLD can quickly converge to within 0.1° in 3 min, and the estimation accuracy and stability are better than the other two methods.

The proposed SALAD is simulated at different latitudes, for example, from -85° to 85°, and each degree is a simulation point. To illustrate the relationship between the local latitude and the latitude and attitude estimation errors, only bias errors of inertial sensors in Table 2 are considered to obtain smooth curves of the latitude and attitude errors by avoiding the effects of random noise. As shown in Fig. 4, whether it is south or north latitude, the latitude error $\delta L$ increases with the decreasing latitude, and when the latitude is close to zero, $\delta L$ will increase rapidly. Note that in Fig. 4, when the local latitude is close to 0°, the sudden decrease of $\delta L$ is caused by the constraint processing of limiting the square root result to less than 1 in (38), which ensures that SALAD is working properly in lower latitude, but it has a small effect on the attitude estimation error around 0°.

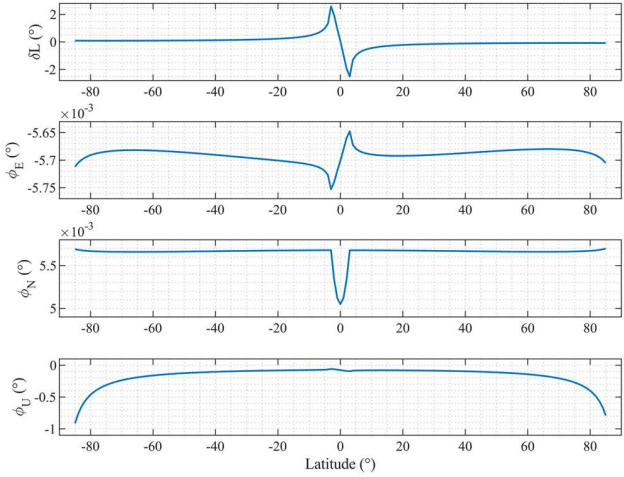

Fig. 4 Error curves of latitude and attitude at different latitudes

The horizontal attitude errors, i.e., $\phi_E$, $\phi_N$, keep in a stable range at different latitudes, and they are close to their alignment limits, which shows $\phi_E$, $\phi_N$ are almost independent of the latitude or slightly affected by the latitude error. For $\phi_U$, the yaw angle error, it is very large in high latitudes, and it decreases slowly with the decreasing latitude, which is because $\phi_U$ is inversely proportional to the cosine of latitude. Generally, the attitude error obtained by SALAD are consistent with the law of the initial alignment error when latitude is known, and $\phi_U$ is less affected by the latitude error, especially in low latitudes. Although the latitude estimation accuracy of SALAD does not meet the positioning requirements, it is sufficient to achieve satisfied initial attitude accuracy for some attitude and heading reference systems. Moreover, even if the latitude accuracy is poor in lower latitudes, it has a small effect on the result of the attitude estimation.

2) Alignment performance of newTRIAD and SALAD

The misalignment angle curves, i.e., $\phi^n = [\phi_E \ \phi_N \ \phi_U]$, are drawn with the alignment time of 180 s for TRIAD, OBA, newTRIAD and SALAD, respectively. An intuitive comparison is presented in Fig. 5 and a root mean square error (RMSE) statistic in the last 30 s is listed in Table 3. In the horizontal direction, the pitch and roll misalignment angles $\phi_E$, $\phi_N$ of the four methods are close to the theoretical alignment limits and the difference among them is very small. Note that the roll misalignment angle of SALAD is slightly larger than the other three methods, this is caused by its latitude estimation error, and it means that $\phi_N$ is sensitive to the latitude. In azimuth direction, the yaw misalignment angles $\phi_U$ of the four methods are close to its theoretical alignment limit as well, but there is a small error of about 0.03 ° due to the random noise of inertial sensors. Moreover, the stability of TRIAD is obviously weaker than that of the other three methods, which reflects the main shortcoming of the dual-vector method compared with the multi-vector methods. However, TRIAD takes the least amount of computing time as shown in Table 3, SALAD spends the most and is more than twice as much as that of TRIAD.

This is because SALAD uses all the observation vectors and needs to compute the sine and cosine functions about the latitude repeatedly in the latitude estimation process, which will increase the computing time. The computing time of the proposed newTRIAD method is less than the classic OBA, which is mainly because the newTRIAD have a low complexity by dealing with the eigenvalue decomposition of two symmetric 3×3 matrices and without iterations in each attitude determination. Note that the statistical time is carried out on a personal computer, and in practice, the commonly used embedded systems are enough for newTRIAD and SALAD.

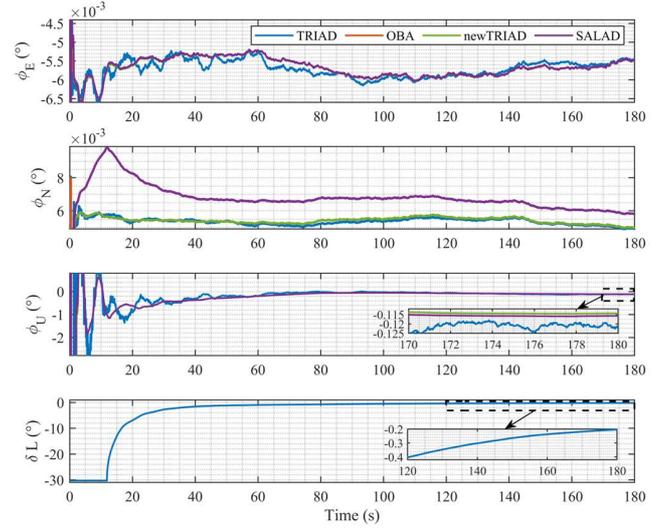

Fig. 5 Curves of the alignment error for four methods. Note that the drawing order of four curves is from the front to the back in the legend, i.e., TRIAD, OBA, newTRIAD, SALAD, and if two curves coincide, the curve drawn first will be covered by the curve drawn later. That is why some curves cannot be seen in the graph. For instance, OBA is covered by newTRIAD in the misalignment angle curves, and the same as following figures.

TABLE 3
RMSEs OF THE ALIGNMENT ERROR DURING 150~180 S

| Method | Pitch (°) | Roll (°) | Yaw (°) | Time (s) [1] |
|---|---|---|---|---|
| TRIAD | 0.005548 | 0.005107 | 0.124047 | 0.1994 |
| OBA | 0.005632 | 0.005163 | 0.110721 | 0.3227 |
| newTRIAD | 0.005632 | 0.005163 | 0.110707 | 0.2651 |
| SALAD [2] | 0.005626 | 0.006025 | 0.112037 | 0.4785 |

[1] The computing time used in the core parts of each method in processing 180 s IMU data.
[2] The latitude RMSE of SALAD is 0.228754°.

As shown in Fig. 5, the misalignment angle curves of the two kinds of multi-vector methods, i.e., OBA and newTRIAD, almost completely coincide, which shows that the proposed newTRIAD is as effective as the popular OBA method. From Table 3, the alignment errors of OBA and newTRIAD are the same, but newTRIAD is about $1.4 \times 10^{-5}$ ° smaller than that of OBA in the yaw misalignment angle.

To verify this slight difference, more simulations are carried out between OBA and newTRIAD methods, including 300 swaying simulations under the condition of randomly generated swaying center with and without inertial sensor errors, respectively. The alignment time lasts for 180 s and no abnormal results found means the newTRIAD can approximate the optimal solution in the whole attitude space, and it is effective for initial alignment. Moreover, the RMS errors of alignment results show that the yaw error of newTRIAD is lower than that of OBA by $10^{-5}$ magnitude with or without sensor errors, and the difference of them in pitch and roll angles is $10^{-8}$ magnitude, which shows that newTRIAD and OBA have very similar alignment performance, even slightly better than OBA in azimuth alignment from a simulation point of view. We guess that this difference comes from the Wahba's loss function, which is based on the method of least squares to achieve an approximate solution of the vehicle attitude by minimize the sum of the squares of the errors from every observation vector. Although OBA could achieve the optimal solution of the Wahba's loss function, but this defined object function maybe not full reflect the true attitude of the vehicle. However, newTRIAD does not depend on the defined object function, it is a completely analytical solution to achieve the real attitude of the vehicle by using as many observation vectors as possible. Note that this is a bold guess, and there is no further theoretical analysis for the above explanation.

*B. Experiment tests*

The experiment tests are carried out by two different self-developed IMUs on a three-axis turntable in our laboratory and a ship in Zhoushan islands, respectively. The IMUs used are consisting of three fiber optic gyroscopes and three quartz accelerometers, and their bias stability are better than 0.03 °/h and 100 μg in turntable tests, and better than 0.005 °/h and 50 μg in ship mooring tests. Moreover, the two IMUs are well calibrated and the output frequency of the IMU is set to 100 Hz. The experiment equipment used is shown in Fig. 6.

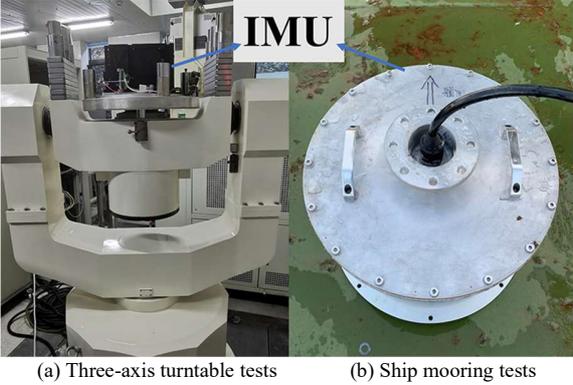

(a) Three-axis turntable tests  (b) Ship mooring tests

Fig. 6 Experiment equipment for alignment tests

1) Quasi-static experiment

The black IMU is mounted on the inner frame of the turntable, and a period of 180 s IMU static data is collected to verify the proposed newTRIAD and SALAD methods. Due to the inconvenient of recording the real-time attitude of the IMU by the turntable in our experience, the reference attitude of the static IMU is provided by the pure inertial navigation results, and the initial attitude of IMU is obtained by the angular position output of the turntable. In this way, both the obtained reference attitude by pure inertial navigation and the attitude alignment results are affected by the IMU error, then the misalignment angle will be not affected to some extent by the IMU error, therefore the calculated misalignment angle is smaller when compared with using a real theoretical attitude. However, this does not affect the verification of the alignment method.

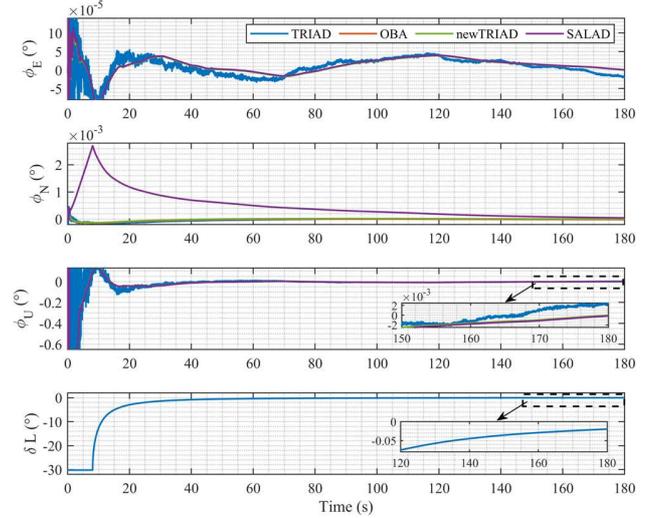

Fig. 7 Curves of the alignment error in the static test

The misalignment angle curves of TRIAD, OBA, newTRIAD, SALAD and the latitude estimation error of MvLD in SALAD are shown in Fig. 7.

Clearly, all the methods can achieve an acceptable alignment accuracy in 60 s and the estimated latitude by SALAD can quickly approach the true value under quasi-static conditions.

2) Turntable swaying experiment

Second, the turntable swaying experiment is performed. The inner, middle, outer frame of the turntable are all swaying in the sine mode with the swaying periods of 6 s, 8 s, 10 s, respectively, and the swaying amplitude of 5 °, and the swaying center attitude is set to [0; 0; 180 °]. Here, the turntable yaw angle is 0 ~ 360 ° in the anticlockwise direction, and the initial attitude is provided by the static turntable attitude, then the reference attitude is obtained by pure inertial navigation and given in Fig. 8. Since the turntable used is a UUT type, the swaying attitude of IMU is not the same as the turntable swaying setting, and it is more complicated. The misalignment angle curves of the four methods are shown in Fig. 9, and the RMSEs of misalignment angles at 471 ~ 500 s are listed in Table 4.

As shown in Fig. 9, the orange curves, i.e., the misalignment angles of OBA, is mostly invisible because it is covered by green curves, which means newTRIAD has a better alignment result as OBA, and in the first 60 s, newTRIAD has a faster convergence speed than OBA and TRIAD in the pitch and yaw alignment. The alignment results of TRIAD are less stable in swaying conditions, and the misalignment angles of SALAD are large due to the poor latitude estimation error in the early stages, then it decreases with the improvement of latitude estimation accuracy. The latitude estimation is heavily affected by gravity observation vectors, and the slower estimation speed in Fig. 9 compared with static conditions is mainly due to the lever arm error, which is clear in Fig. 6 (a) and it is not well compensated, thus alignment results listed in Table 4 is not better than the static test, however, the proposed newTRIAD is still slightly better than OBA and TRIAD.

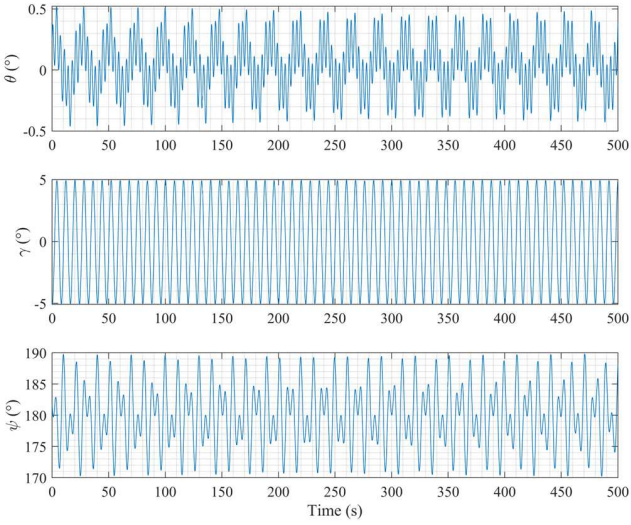

Fig. 8 Attitude reference curves in turntable swaying tests

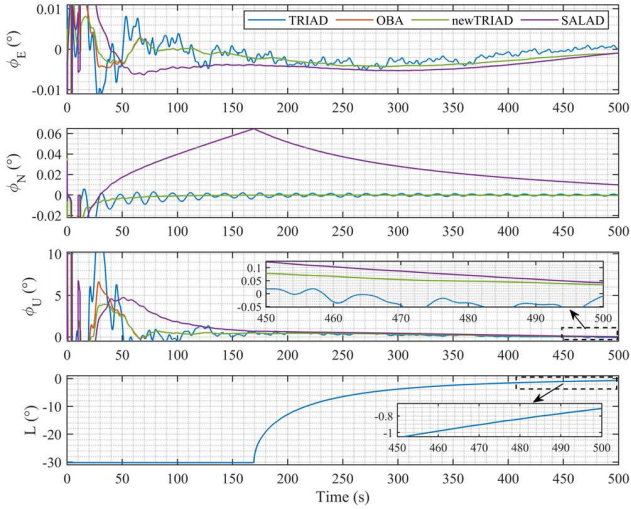

Fig. 9 Curves of the alignment error in the turntable swaying test

TABLE 4
RMSEs OF THE ALIGNMENT ERROR IN THE TURNTABLE TEST OF 471~500 s

| Method | Pitch (°) | Roll (°) | Yaw (°) |
|---|---|---|---|
| TRIAD | $5.60904\times10^{-4}$ | $5.98628\times10^{-4}$ | $4.75170\times10^{-2}$ |
| OBA | $1.13620\times10^{-3}$ | $4.23371\times10^{-4}$ | $4.50483\times10^{-2}$ |
| newTRIAD | $1.13555\times10^{-3}$ | $4.05785\times10^{-4}$ | $4.50245\times10^{-2}$ |
| SALAD | $1.36482\times10^{-3}$ | $1.07660\times10^{-2}$ | $6.45483\times10^{-2}$ |

Note that the latitude RMS error of SALAD is 0.80597°.

3) Ship mooring experiment

Last, ship mooring experiment in Zhoushan islands is used to further verify the validity of newTRIAD and SALAD. Fig. 10 is the reference attitude of the white IMU on the moored ship, and it is also provided by the pure inertial navigation. Clearly, the swaying amplitude and period of pitch, roll, yaw angles increase in order, and it is a normal ship mooring condition. The alignment error is shown in Fig. 11, and the RMS error is listed in Table 5. Generally, the trend of the misalignment angle curves is like that of Fig. 9. For instance, a large change of the yaw angle in 30 ~ 50 s has a greater impact on four alignment methods, but newTRIAD is affected by a smaller fluctuation than OBA and TRIAD and has a faster convergence speed, and the alignment result of newTRIAD is more stable than that of TRIAD, even with higher wiggle frequencies, irregular wobbles, and slightly linear vibrations, this is mainly due to newTRIAD utilizing more observation vectors than TRIAD. Compared with OBA, newTRIAD has similar alignment accuracy with OBA, and slightly better than OBA in 271 ~ 300 s, which is consistent with the results of the simulation. Moreover, the higher alignment accuracy of newTRIAD and OBA in Table 5 indicates that GAM-based alignment methods is more suitable for high-precision IMUs.

For SALAD, it mainly depends on the estimation accuracy of latitude, and as the latitude error decreases, it can eventually converge to an acceptable alignment accuracy of 0.57127 ° by increasing the latitude estimation time to 900 s, thus it is necessary to further analyze how to improve the accuracy and speed of latitude estimation. As shown in Fig. 5, Fig. 7, Fig. 9, and Fig. 11, the latitude error curves of SALAD are smooth and monotonous close to zero, which is different from other methods. Take Fig. 5 as an example, the latitude estimation process can be roughly divided into three stages, and it is mainly related to the measurement error of gravity observation vectors. In the early stage of latitude estimation, fewer observation vectors are used, and the total measurement error of observation vectors is very large, which makes the latitude determination, i.e. (38), invalid by the violation of the arc-cosine function constraint. Owing to the constraint processing, the latitude is set to zero for the next update, and this interval (0~12 s) is defined as the growing stage. As time increases, more gravity observation vectors are involved in latitude estimation, then the total measurement error of observation vectors is smoothed gradually. Once the error is less than a certain threshold, equation (38) starts to work and the estimated latitude is rapidly approaching the optimal solution, and this interval (13~50 s) is defined as the accelerating stage. When the error becomes smaller and maintains in a certain range, the estimated latitude is close to the optimal solution, and this interval (51~180 s) is defined as the stable stage. However, for the experiment result shown in Fig. 11, the latitude estimation time of the three stages is much longer than that of simulation, it is mainly due to the measurement error of gravity observation vectors, which is derived from the linear vibration error, the lever arm error, etc., whereas these error sources are simplified or omitted in the simulation. Therefore, if SALAD includes the compensation of the lever arm and linear vibration errors,

TABLE 5
RMSEs OF THE ALIGNMENT ERROR IN SHIP MOORING TEST

| Method | Pitch (°) | | Roll (°) | | Yaw (°) | |
|---|---|---|---|---|---|---|
| | 271-300 s | 471-500 s | 271-300 s | 471-500 s | 271-300 s | 471-500 s |
| TRIAD | $1.36810\times10^{-4}$ | $7.73992\times10^{-5}$ | $7.52457\times10^{-4}$ | $2.92396\times10^{-4}$ | $1.22555\times10^{-2}$ | $4.21655\times10^{-3}$ |
| OBA | $2.48689\times10^{-5}$ | $8.21761\times10^{-5}$ | $2.06808\times10^{-4}$ | $1.98045\times10^{-5}$ | $1.90047\times10^{-3}$ | $3.62749\times10^{-4}$ |
| newTRIAD | $2.24647\times10^{-5}$ | $1.02915\times10^{-5}$ | $2.24866\times10^{-4}$ | $7.68627\times10^{-6}$ | $1.68272\times10^{-3}$ | $4.77172\times10^{-4}$ |
| SALAD | $3.51937\times10^{-4}$ | $3.35849\times10^{-4}$ | $3.20513\times10^{-2}$ | $2.00629\times10^{-2}$ | $5.62760\times10^{-3}$ | $3.18206\times10^{-2}$ |

The latitude RMS error of SALAD is 5.35442° in 271~300 s, 1.87189 ° in 471~500 s, and 0.57127 ° in 871~900 s.

and the IMU error estimation, it can effectively improve the speed and accuracy of the latitude estimation in actual applications.

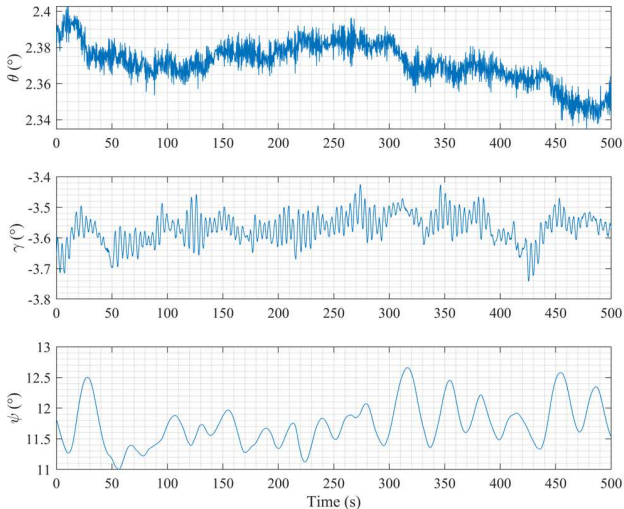

Fig. 10 Attitude reference curves in the ship mooring test

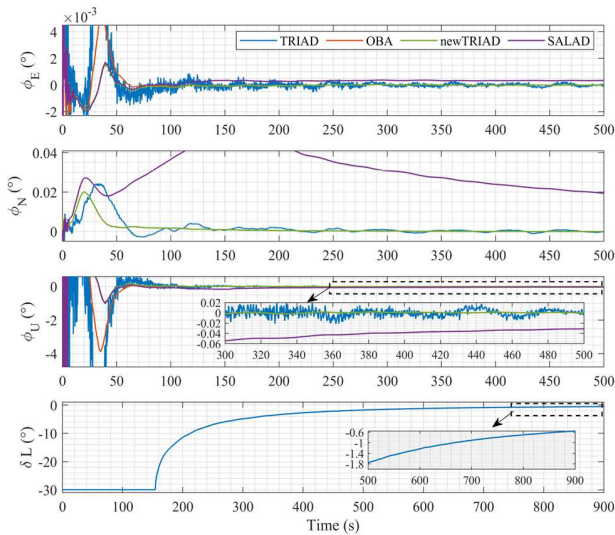

Fig. 11 Curves of the alignment error in the ship mooring test

## IV. CONCLUSION

This paper presents a new type of initial alignment method under swaying and latitude uncertain conditions, and its main feature is utilizing all observation vectors by constructing the dyadic tensor of the observation vector itself without using the defined Wahba's loss function. This feature can be used to estimate the optimal initial attitude, like newTRIAD, which is completely different from the common way of estimating attitude by solving the Wahba problem, such as OBA, etc., and it can be used to estimate the local latitude to realize self-alignment, like SALAD, which is completely different from the existing latitude estimation method through the geometric relation of limited observation vectors, and the latitude and attitude estimation of SALAD share the same calculation process, which can improve the alignment efficiency and is different from the common method that the latitude and attitude estimation processes are independent of each other. Generally, newTRIAD and SALAD are not only simple in terms of conceptual and computational complexity, but also can utilize all observation vectors. Since the calculation mainly involves the construction of symmetric 3×3 dyadic tensors that can be easily diagonalized to finish their eigenvalue decomposition, it is sufficient for most embedded systems.

Simulation and experiment tests have verified the effectiveness of newTRIAD and SALAD. Compared with TRIAD, OBA methods, newTRIAD not only has the advantages of faster convergence speed and higher stability than TRIAD, but also has very similar alignment accuracy with OBA, and it is even better than OBA in azimuth alignment. SALAD is a completely self-alignment method, which can achieve a real-time estimation of latitude and attitude and provide similar alignment accuracy to newTRIAD when the latitude is not known. Moreover, the positive-negative sign of the latitude can also be determined by SALAD, and the estimated accuracy of latitude is improved by using all observation vectors. In practical applications, the error compensation and noise processing of IMU are beneficial for SALAD to further improve the estimation speed and accuracy of latitude and attitude, and some modified SALAD can be easily applied to in-motion latitude estimation and attitude alignment, which will be studied in the future. In general, this study presents a new way for initial alignment based on all observation vectors, and the proposed latitude and attitude estimation method is a starting attempt, which may be of great potential for the improvement in the field of vehicle attitude references and navigation, such as the swaying alignment of the moored ship, the self-alignment of underwater and underground vehicles, the in-motion alignment with a velocity-aided, the latitude revision and heading determination of shipborne gyrocompass system, etc.

## REFERENCES

[1] J. Dunik, S. K. Biswas, A. G. Dempster, T. Pany, and P. Closas, "State Estimation Methods in Navigation: Overview and Application," *IEEE Aerosp. Electron. Syst. Mag.*, vol. 35, no. 12, pp. 16–31, 2020, doi: 10.1109/MAES.2020.3002001.

[2] Y. Ben, X. Zang, Q. Li, and J. He, "A dual-state filter for a relative velocity aiding Strapdown Inertial Navigation System," *IEEE Trans. Instrum. Meas.*, vol. 9456, no. c, pp. 1–1, 2020, doi: 10.1109/tim.2020.3010042.

[3] J. Li, W. Gao, Y. Zhang, and Z. Wang, "Gradient Descent Optimization-Based Self-Alignment Method for Stationary SINS," *IEEE Trans. Instrum. Meas.*, vol. PP, pp. 1–9, 2018, doi: 10.1109/TIM.2018.2878071.

[4] S. Han, X. Ren, J. Lu, and J. Dong, "An Orientation Navigation Approach Based on INS and Odometer Integration for Underground Unmanned Excavating Machine," *IEEE Trans. Veh. Technol.*, vol. 69, no. 10, pp. 10772–10786, 2020, doi: 10.1109/TVT.2020.3010979.

[5] W. Li, W. Wu, J. Wang, and L. Lu, "A fast SINS initial alignment scheme for underwater vehicle applications," *J. Navig.*, vol. 66, no. 2, pp. 181–198, 2013, doi: 10.1017/S0373463312000318.

[6] S. Tanygin and M. D. Shuster, "The many TRIAD algorithms," *Adv. Astronaut. Sci.*, vol. 127 PART 1, pp. 81–100, 2007.

[7] J. Li, Y. Li, and B. Liuxs, "Fast Fine Initial Self-alignment of INS in Erecting Process on Stationary Base," *J. Navig.*, vol. 71, no. 3, pp. 697–710, 2018, doi: 10.1017/S037346331700090X.

[8] W. Gao, B. Lu, and C. Yu, "Forward and backward processes for INS compass alignment," *Ocean Eng.*, vol. 98, pp. 1–9, 2015, doi: 10.1016/j.oceaneng.2015.01.016.

[9] F. Pei, S. Yin, and S. Yang, "Rapid Initial Self-Alignment Method Using CMKF for SINS under Marine Mooring Conditions," *IEEE Sens. J.*, vol. 21, no. 8, pp. 9969–9982, 2021, doi: 10.1109/JSEN.2021.3055536.

[10] M. Wu, Y. Wu, X. Hu, and D. Hu, "Optimization-based alignment for inertial navigation systems: Theory and algorithm," *Aerosp. Sci. Technol.*, vol. 15, no. 1, pp. 1–17, 2011, doi: 10.1016/j.ast.2010.05.004.

[11] X. Liu, X. Xu, Y. Zhao, L. Wang, and Y. Liu, "An initial alignment method for strapdown gyrocompass based on gravitational apparent motion in inertial frame," *Meas. J. Int. Meas. Confed.*, vol. 55, pp. 593–604, 2014, doi: 10.1016/j.measurement.2014.06.004.

[12] X. Xu, X. Xu, T. Zhang, Y. Li, and J. Tong, "A kalman filter for sins self-alignment based on vector observation," *Sensors (Switzerland)*, vol. 17, no. 2, pp. 1–19, 2017, doi: 10.3390/s17020264.

[13] Y. Huang, Y. Zhang, and X. Wang, "Kalman-Filtering-Based In-Motion Coarse Alignment for Odometer-Aided SINS," *IEEE Trans. Instrum. Meas.*, vol. 66, no. 12, pp. 3364–3377, 2017, doi: 10.1109/TIM.2017.2737840.

[14] T. Gaiffe, Y. Cottreau, N. Faussot, G. Hardy, P. Simonpietri, and H. Arditty, "Highly compact fiber optic gyrocompass for applications at depths up to 3000 meters," *Proc. 2000 Int. Symp. Underw. Technol. UT 2000*, pp. 155–160, 2000, doi: 10.1109/UT.2000.852533.

[15] H. A. Hashim, "Attitude determination and estimation using vector observations: Review, challenges and comparative results," *arXiv*, 2020.

[16] H. D. Black, "A passive system for determining the attitude of a satellite," *AIAA J.*, vol. 2, no. 7, pp. 1350–1351, 1964, doi: 10.2514/3.2555.

[17] G. Wahba, "A Least Squares Estimate of Satellite Attitude," *SIAM Rev.*, vol. 7, no. 3, pp. 409–409, Jul. 1965, doi: 10.1137/1007077.

[18] F. L. Markley and D. Mortari, "Quaternion attitude estimation using vector observations," *J. Astronaut. Sci.*, vol. 48, no. 2–3, pp. 359–380, 2000, doi: 10.1007/bf03546284.

[19] J. Wu, Z. Zhou, B. Gao, R. Li, Y. Cheng, and H. Fourati, "Fast Linear Quaternion Attitude Estimator Using Vector Observations," *IEEE Trans. Autom. Sci. Eng.*, vol. 15, no. 1, pp. 307–319, 2018, doi: 10.1109/TASE.2017.2699221.

[20] R. P. Patera, "Attitude estimation based on observation vector inertia," *Adv. Sp. Res.*, vol. 62, no. 2, pp. 383–397, Jul. 2018, doi: 10.1016/j.asr.2018.04.039.

[21] Q. Fu, F. Wu, S. Li, and Y. Liu, "In-Motion Alignment for a Velocity-Aided SINS with Latitude Uncertainty," *IEEE/ASME Trans. Mechatronics*, vol. 25, no. 6, pp. 2893–2903, 2020, doi: 10.1109/TMECH.2020.2997238.

[22] G. M. Yan, W. S. Yan, D. M. Xu, and H. Jiang, "SINS initial alignment analysis under geographic latitude uncertainty," *Aerosp. Control*, vol. 26, no. 2, pp. 31–34, 2008.

[23] S. Wang, G. Yang, W. Chen, and L. Wang, "Latitude determination and error analysis for stationary SINS in unknow-position condition," *Sensors (Switzerland)*, vol. 20, no. 9, pp. 1–18, 2020, doi: 10.3390/s20092558.

[24] Y. Wang, J. Yang, and B. Yang, "SINS initial alignment of swaying base under geographic latitude uncertainty," *Hangkong Xuebao/Acta Aeronaut. Astronaut. Sin.*, vol. 33, no. 12, pp. 2322–2329, 2012.

[25] X. Liu, X. Liu, Q. Song, Y. Yang, Y. Liu, and L. Wang, "A novel self-alignment method for SINS based on three vectors of gravitational apparent motion in inertial frame," *Meas. J. Int. Meas. Confed.*, vol. 62, pp. 47–62, 2015, doi: 10.1016/j.measurement.2014.11.010.

[26] J. Li, W. Gao, and Y. Zhang, "Gravitational Apparent Motion-Based SINS Self-Alignment Method for Underwater Vehicles," *IEEE Trans. Veh. Technol.*, vol. 67, no. 12, pp. 11402–11410, 2018, doi: 10.1109/TVT.2018.2876469.

[27] J. Wang, T. Zhang, J. Tong, and Y. Li, "A Fast SINS Self-Alignment Method under Geographic Latitude Uncertainty," *IEEE Sens. J.*, vol. 20, no. 6, pp. 2885–2894, 2020, doi: 10.1109/JSEN.2019.2957839.

[28] G. Yan, S. Li, W. Gao, J. Li, and J. Ren, "An improvement for SINS anti-rocking alignment under geographic latitude uncertainty," *Zhongguo Guanxing Jishu Xuebao/Journal Chinese Inert. Technol.*, vol. 28, no. 2, pp. 141–146, 2020, doi: 10.13695/j.cnki.12-1222/o3.2020.02.001.

[29] Y. Liu, X. Xu, X. Liu, Y. Yao, L. Wu, and J. Sun, "A self-alignment algorithm for SINS based on gravitational apparent motion and sensor data denoising," *Sensors (Switzerland)*, vol. 15, no. 5, pp. 9827–9853, 2015, doi: 10.3390/s150509827.

[30] P. M. G. Silson, "Coarse alignment of a ship's strapdown inertial attitude reference system using velocity loci," *IEEE Trans. Instrum. Meas.*, vol. 60, no. 6, pp. 1930–1941, 2011, doi: 10.1109/TIM.2011.2113131.